\documentclass[letterpaper]{article} 
\usepackage[preprint]{aaai2027} 
\usepackage[hyphens]{url}  
\usepackage{graphicx} 
\usepackage{natbib}  
\usepackage{caption} 
\usepackage{booktabs}
\usepackage{xcolor}

\newcommand{\stmt}[1]{\textit{#1}}
\newcommand{\releasenote}{This paper is under review as a workshop paper. We will release the repository upon acceptance.}
\title{Disclosure Bias in AI Reviewers}

\author{
    Jens Leysen\textsuperscript{\rm 1},
    Noah Dani\"{e}ls\textsuperscript{\rm 1},
    Marco Favier\textsuperscript{\rm 1},
    Bart Goethals\textsuperscript{\rm 1,\rm 2}
}
\affiliations{
    \textsuperscript{\rm 1}University of Antwerp, Antwerp, Belgium\\
    \textsuperscript{\rm 2}Froomle, Antwerp, Belgium
}

\begin{document}

\maketitle

\begin{abstract}
AI models increasingly help to review papers, and many conferences and journals now require authors to state how they used AI. Does this AI-use statement change how an AI reviewer rates a paper, and could it change which papers are accepted? In a preliminary study, we varied only the AI-use statement of 150 ICLR 2027 submissions, in up to nine versions adapted from real statements: from no AI, through writing help and coding, to an AI research agent and an AI author. Six AI reviewers rated the versions: the decision model Jev and the language models Claude Haiku~5.5, Claude Sonnet~5.5, Gemini~3.8 Flash, Gemini~3.1 Pro and Gemini~4 Argon. Across 23{,}652 reviews, 
we find that every reviewer rates papers lower on average when the statement says AI did all the work than under a statement of no AI use. The AI research agent statement lowers the rating significantly for Jev, Sonnet and Argon.
We find no evidence of a disclosure bias for AI assistance: for writing help, coding and research help, no reviewer's rating differs significantly from its rating under a statement of no AI use. 
Whether the observed penalty changes which papers are accepted depends on more than the size of the penalty: on how finely and how consistently the reviewer separates the papers near the acceptance boundary, on whether the penalty is spread over all papers or concentrated on a few, and possibly on how many other papers disclose AI use. 
\end{abstract}

\section{Introduction}
\label{sec:introduction}

Machine learning venues now require authors to say how they used AI. From 2027, every ICLR submission must contain a section stating it, listing tasks such as implementing methods, interpreting results or assisting with proofs~\cite{iclr2027aipolicy}. At the same time, AI models are entering the reviewing process: a growing share of review text at AI conferences may already be substantially modified by LLMs~\cite{liang2024monitoring}, LLM-assisted reviews give different scores~\cite{latona2024lottery}, ICLR itself has sent LLM feedback to its reviewers~\cite{thakkar2025iclr}, and AAAI-26 showed an AI-generated review of every fully reviewed paper to the committee members deciding on it~\cite{biswas2026aaai}. A paper's AI-use statement will therefore increasingly be read by an AI.

Disclosure policies rest on the assumption that an honest statement does not hurt a paper that is otherwise sound. There is evidence that people place less trust in those who disclose AI use~\cite{schilke2025transparency}. LLM judges, in turn, respond to labels about who wrote other kinds of text~\cite{cheong2025penalizing,saraf2025label,chae2026labels}. 
Whether the assumption holds for AI reviewers is unknown: to the best of our knowledge, no study has asked how an AI reviewer reads the AI-use statement of a scientific paper, whether its reading depends on how much the AI did, or whether it changes which papers are accepted. Yet the answer decides whether a disclosure policy rewards honesty or under-reporting.

We ask whether an AI-use statement changes an AI reviewer's rating of a paper, and discuss when such a change would alter which papers are accepted. We call a change in rating caused by the statement alone a \emph{disclosure bias}. We answer the first question with a controlled experiment. We took 150 real ICLR 2027 submissions and replaced only the body of their AI-use section. In this preliminary study, we use one statement per level of AI use, each close to real statements in our sample, from writing help to an AI research agent, and an AI author as an extreme. Six AI reviewers of different kinds then filled in the ICLR review form for the paper versions. One is Jev, a decision model that answers with probabilities~\cite{typesafe2026docs}. The other five are language models: Claude Haiku~5.5 and Claude Sonnet~5.5, and Gemini~3.8 Flash, Gemini~3.1 Pro and Gemini~4 Argon. We ask each language model either for a rating alone or for a written review first. Since every contrast compares identical papers, any difference beyond the noise of re-reviewing comes from the AI-use statement. We find:

\begin{itemize}
\item \textbf{No evidence of a disclosure bias for AI assistance.} For writing help, AI-written code and AI research help, no reviewer's rating differs significantly from its rating under a statement of no AI use. 
\item \textbf{Some evidence for a disclosure bias against AI agents and authors.} Every reviewer, by contrast, rates papers lower on average when the statement says AI did all the work; an AI research agent statement lowers the rating significantly for Jev, Sonnet and Argon.
\item \textbf{The rating penalty alone does not show how many decisions change.} In a simulation where a single AI reviewer decides and only one paper discloses agentic use, the statistically significant penalties reject 8 to 18 percentage points more of the accepted papers than re-reviewing alone. On Jev's hypothetical expected rating, which is far finer, the same statement rejects 77 percentage points more.
\end{itemize}

\section{Related Work}
\label{sec:related}

\paragraph{LLMs in peer review.} LLM feedback on papers overlaps with human reviews about as much as human reviews overlap with each other~\cite{liang2023feedback}. An estimated 6.5--16.9\% of review text at four AI conferences, ICLR among them, could have been substantially modified by an LLM~\cite{liang2024monitoring}; at ICLR 2024, AI-assisted reviews gave higher scores, and borderline papers with such reviews were more likely to be accepted~\cite{latona2024lottery}. Venues now also build AI into their reviewing process: at ICLR 2025, an LLM gave reviewers feedback on more than 20{,}000 of their reviews in a randomised study~\cite{thakkar2025iclr}, ICML 2026 compared reviewer policies on LLM use in a randomised experiment~\cite{kim2026icml}, and AAAI-26 attached an AI-generated review to each of its 22{,}977 fully reviewed papers, visible to the senior program committee members and area chairs who decided on these papers~\cite{biswas2026aaai}.

\paragraph{Cues that sway LLM reviewers.} LLM reviewers respond to content other than the science. Instructions hidden in a manuscript can raise an LLM reviewer's rating~\cite{ye2024risks}, and some authors have hidden such instructions in their preprints~\cite{lin2026hidden}. LLM reviews also take up the limitations that authors point out far more often than human reviews do~\cite{ye2024risks}. Identical manuscripts are rated higher when their authors come from prestigious institutions~\cite{howell2025prestige,macharlavasu2026justice}, although not in every audit study~\cite{alharbi2026scrutinise}. LLM reviewers rate LLM-written papers higher than human-written ones~\cite{li2025reval}, and LLMs prefer LLM-written descriptions of the same items~\cite{laurito2025aiai}; in real reviews, the apparent leniency of LLM-assisted reviews towards LLM-assisted papers disappears once paper quality is accounted for~\cite{sharma2026favor}. LLM judges also favour their own outputs~\cite{panickssery2024recognize}, a preference that may stem from the familiarity of the text~\cite{wataoka2024selfpreference}.

\paragraph{Labels without a change in the text.} A label alone can move LLM judges. Responses labelled as written by an expert or a human are preferred over those labelled as written by an LLM, without the label appearing in the judges' justifications~\cite{marioriyad2025silent}. In one study, naming the model that wrote a text shifted scores, up for Claude and down for Gemini, whichever model wrote it~\cite{saraf2025label}. In another study, merely telling a judge that a research proposal was its own or another model's shifted its ratings~\cite{chae2026labels}. Closest to our question, a line stating that a news article was created with the help of AI tools lowered its ratings slightly, by human readers and by two LLMs alike~\cite{cheong2025penalizing}.

\paragraph{AI-use disclosure.} Most top journals had guidance on generative AI, often with specific disclosure requirements~\cite{ganjavi2024publishers}; ICLR requires authors to disclose any use of LLMs~\cite{iclr2025llmpolicy}, from 2027 in a mandatory section~\cite{iclr2027aipolicy}. At ICLR 2026, 64\% of submissions included such a statement, nearly all of them mentioning editing or polishing~\cite{mohapatra2026disclosure}. Elsewhere, disclosure lags far behind actual use: in journals in early 2025, about 40 papers showed statistical signs of AI use for every one that disclosed it~\cite{he2025policies}. Disclosure can also carry a cost. People who disclose AI use are trusted less~\cite{schilke2025transparency}, and people known to use AI are seen as less competent~\cite{reif2025penalty}. The penalty grows with the AI's role: readers discount AI-generated writing more than AI-edited writing~\cite{li2024disclosure}, and an ``AI-generated'' label hurts mainly when it is read as full automation~\cite{altay2024headlines}. One study of ML experts, however, found no such penalty for abstracts. They rated three versions of a paper's abstract: written by people, generated by an LLM, or edited by one. Readers who were told which version was which rated all three higher than readers who had to guess, and chose the LLM-edited version most often~\cite{akpinar2026summaries}. For reviewers, the evidence is thinner. In a study of scientific abstracts, accept decisions did not depend on whether the text was AI-generated~\cite{hazra2026accepted}. In real peer review, human reviewers rate weak papers lower when they contain detected LLM text, while strong papers are not affected~\cite{sharma2026favor}. 

To our knowledge, no study has tested whether an AI-use statement alone changes a human reviewer's rating. Nor has any study asked how an AI reviewer responds to a paper's AI-use statement. The label studies closest to ours change a label on a generic text. Most name the model or the kind of author that wrote it. One adds a line saying that a news article was written with the help of AI tools~\cite{cheong2025penalizing}. The audit studies of LLM reviewers vary the authors and their affiliations, not the statement. The studies of how much the AI did concern human readers, mostly of news and essays. We test the AI-use statements that venues now require, on real submissions. We vary only what the statement says the AI did and which model it names, and six AI reviewers fill in a conference review form. We then ask whether the statement alone could change which papers are accepted.

\section{Method}
\label{sec:method}
\begin{table*}[t]
\caption{Our statements: one for each level of AI use found in the sample, plus \stmt{No AI} and \stmt{AI author}. Three variants of the \stmt{Agent} statement make nine statements in all: \stmt{Agent (ChatGPT)}, \stmt{Agent (Claude)} and \stmt{Agent (Gemini)} name their respective model instead of ``generative AI tools''.}
\label{tab:ladder}
\centering
\small
\begin{tabular}{lp{0.8\textwidth}}
\toprule
Level & Statement \\
\midrule
0 \stmt{No AI} & No generative AI tools were used in this work. \vspace*{3pt}\\
1 \stmt{Writing} & We used generative AI tools solely for grammar checking and polishing the writing. We did not use them for research ideation, experimental design, code development or data analysis. The authors take responsibility for the content. \vspace*{3pt}\\
2 \stmt{Code} & We used generative AI tools to implement and refactor the experimental code and to edit the manuscript. The research ideas and experimental design are the authors' own. All generated code was reviewed and run by the authors. The authors take responsibility for the content. \vspace*{3pt}\\
3 \stmt{Research help} & We used generative AI tools to give feedback on the methodology and experimental design, implement methods, analyze and interpret the results, and edit the manuscript. The authors reviewed and verified all AI-assisted work. The authors take responsibility for the content. \vspace*{3pt}\\
4 \stmt{Agent} & This paper results from extended work with generative AI tools as a research agent: they planned and ran the experiments, analyzed the results, and drafted the main text and figures, with the authors reviewing each step. The authors take responsibility for the content. \vspace*{3pt}\\
5 \stmt{AI author} & This paper was written entirely by generative AI tools: they developed the research ideas, wrote the code, ran the experiments, analyzed the results and wrote the text. \\
\bottomrule
\end{tabular}
\end{table*}

Our code, prompts, statements and run configurations, including seeds, will be released in an open repository.\footnote{\releasenote}

\subsection{Papers}
We drew ICLR 2027 submissions at random from OpenReview, all with an AI-use section, which ICLR 2027 makes mandatory. In each version, we keep the section's heading and replace only its body, in place and in the style of the body text. Each of our statements must fit in the space of the paper's own section, so that the rest of the page stays in place. Of the 160 papers we drew, 10 (6\%) had a section too short for our longest statement, so we skipped them and kept 150. The 150 papers we reviewed span 19 of ICLR's primary areas, and the submissions are publicly available on OpenReview. We ran the reviews between 23 October and 5 November 2026, while the submissions were under review at ICLR. We do not name the submissions or publish our altered versions, and we ran all AI reviewers under settings and terms that exclude the papers from being used for model training.

\subsection{Statements}
We read the AI-use statements of all 160 papers we drew. One reports AI only as part of its method; we grouped the other 159 into four levels by what the AI did at most: \stmt{Writing}, \stmt{Code}, \stmt{Research help} and \stmt{Agent}. For each level, our statement closely follows one or a few real statements from that level, shortened and with the same closing sentence on responsibility (Table~\ref{tab:ladder}). Real statements within a level differ widely in wording and in what they report. Each level is therefore represented by one of many possible wordings, and a reviewer may react to that wording as much as to the level. 

The levels differ in how much of the research the AI did itself. The quoted phrases below are taken from the papers' own statements. \stmt{Writing} (61 papers) covers help with the text, such as ``light language editing, readability polishing'' or ``grammatical correction and rephrasing'', and also translation, literature search, formatting and figures. \stmt{Code} (37) adds help with code, such as ``debugging code''. \stmt{Research help} (37) adds input that the authors then act on: ideas, analysis, interpretation of results, ``feedback on conceptual framing, experimental design'' or help ``in the writing of proofs''. \stmt{Agent} (24) means that the AI did research work largely on its own, such as running experiments or synthesising the literature: ``setting up experiments, inspecting results'' or ``research ideation and execution''. None reports using no AI.

Our \stmt{Writing}, \stmt{Code} and \stmt{Research help} statements read much like real ones. Our \stmt{Agent} statement is more explicit than most real ones in two ways. First, most real statements present the AI as helping the authors. They say that AI ``assisted'', and some that mention experiments say that it ``helped'' run them. Such wording fits any share of the work, just as a co-author who largely ran the experiments independently can be said to have helped run them. Ours says plainly that the AI planned and ran the experiments. A few real statements are as plain as ours. In one, an AI assistant ``executed the runs on the authors' hardware''. In another, the steps of an agent team, alternating with manual review, ``span experiment planning and execution, main text and figure generation''. Second, they usually report one or two agentic tasks, while ours lists them all: planning and running the experiments, analysing the results and drafting the text. Other parts of our statements are common in real ones. Most real statements name no model and speak of ``generative AI tools''. Most say that the authors reviewed or verified the AI's work, and most say that the authors take responsibility for the content. Our statements do the same. Since a named model may be read differently, the \stmt{Agent} statement also appears with ChatGPT, Claude or Gemini in place of the generic tools. \stmt{No AI} and \stmt{AI author} have no counterpart among the 160 statements. \stmt{No AI} is the baseline. \stmt{AI author} is a positive control: an extreme case, but one we expect any reviewer who reads the statement to react to.

Some real statements are vague about what the AI did. Some name a task without saying what it covers. ``Assist with writing this paper'' may mean polishing the text, but a written paper is also the end product of the research, so it may also mean help with the research itself. ``Data generation'' does not say which data the AI generated, or for what. Other statements name a broad part of the research, such as ``research development'' or ``research discussion'', without saying which tasks the AI took on. We placed such statements at the lowest level they could mean.

\subsection{Review form}
All reviewers answer the same review form, based on ICLR 2026's. It keeps only the scores and the overall rating, in more compact wording, and asks whether to accept the paper (yes or no). Soundness, presentation and contribution each have four levels: poor, fair, good and excellent (ICLR's 1 to 4). The overall rating has six levels: strong reject, reject, borderline reject, weak accept, accept and strong accept (ICLR's 0 to 10, in steps of two). 

\subsection{Reviewers}
\begin{itemize}
\item \textbf{Jev} (version 1.13.0, TypeSafe AI) is a decision model: rather than generating text, it returns a probability for each answer to a typed question~\cite{typesafe2026docs}. It receives the paper's text up to the references. We read Jev's answer in two ways. Its most likely level, the mode of this distribution, is one of the form's six levels, a rating as a reviewer would give it. Its expected level, the average of the levels weighted by their probabilities, can fall between two levels. No review form allows such a rating; we report it as a hypothetical, finer reading, which the Results use to show what such a reading changes. Both come from the same reviews, so they are two readings of one result, not two independent results.

\item \textbf{Claude Haiku~5.5 and Gemini~3.8 Flash:} These are small general-purpose language models, called Haiku and Flash below. Each receives the paper as a PDF. Because a model may judge differently once it has reasoned about a paper, we use each in two settings, both at low reasoning effort: a \emph{rating} setting, in which we ask it to answer the form directly, without writing a review, and a \emph{review} setting, in which we ask it to first write a review (a summary, strengths, weaknesses and questions for the authors) and then answer the form, in the same response.

\item \textbf{Claude Sonnet~5.5, Gemini~3.1 Pro and Gemini~4 Argon:}
We also test larger models from each provider, \emph{Sonnet} and \emph{Pro}, and Google's newest, \emph{Argon}, to see whether the results hold across models. For reasons of cost, Sonnet and Argon review the papers only under four statements (\stmt{No AI}, \stmt{Writing}, \stmt{Agent} and \stmt{AI author}), in both \emph{review} and \emph{rating} settings. 
\end{itemize}
When asked for a rating alone, the language models barely reason at low effort: the median is 0 reasoning tokens for all but Pro (166). Before a written review, Haiku reasons more (median about 2{,}100 tokens) and Pro about 400; the others still use a median of 0.
We left temperature at each model's default.
Each reviewer gave two independent reviews of every version assigned to it.\footnote{Except three papers for Argon: Google's safety filter blocked every Argon review of them, citing an image in the PDF, so that Argon reviewed 147 papers and we report 23{,}652 reviews instead of 23{,}700. Flash and Pro reviewed the same three papers without a problem.}

\subsection{Rating comparisons}

For each reviewer and statement, we average a paper's two reviews and compare the result with the same paper under the \stmt{No AI} statement. We report the mean difference over the 150 papers. We test each difference with a two-sided paired $t$-test over the papers, which also gives its 95\% confidence interval. A comparison, such as \stmt{Agent} against \stmt{No AI}, is tested for every reviewer and setting; we treat these tests as one family and correct them with Holm's method~\cite{holm1979simple}, which keeps the chance of any false positive within the family at or below 5\%. We call a result significant only if it passes this correction.

\subsection{Simulated decisions}
A rating penalty matters to authors if it changes which papers are accepted. To see when it does, we simulate a simple venue. A single AI reviewer rates every submission, and the venue keeps the best-rated 27.4\%, ICLR 2026's acceptance rate~\cite{iclr2026retrospective}. A paper is thus accepted for its rank among the submissions, not for its rating alone. Real venues decide differently, with several reviewers and an area chair per paper. The simulation isolates what one AI reviewer would decide on its own.

Each paper version was reviewed twice. All papers are first assigned the \stmt{No AI} statement, and one of their two \stmt{No AI} reviews decides which are accepted. 
This review gives the reference ranking of the papers. 
We then let one accepted paper at a time disclose agent use: it gets the \stmt{Agent} statement, while all other papers keep their reference review. The paper is now judged by one of the two reviews of its \stmt{Agent} version. It is rejected if that review ranks it below the best rejected paper. 

A new review can also reject a paper without any change in the statement, because no reviewer is fully consistent. We therefore also judge each accepted paper by its other \stmt{No AI} review, as a control: it shows how many accepted papers re-reviewing alone would reject. The statement's effect is the share rejected beyond this control. Ties in rating are broken at random, by the same rule for the \stmt{Agent} review and the control, and we average over many random draws. 

We run the simulation twice, exchanging the roles of the two reviews of each version, and pool both runs. We report the effect with a bootstrap 95\% confidence interval and a two-sided permutation test, Holm-corrected across reviewers and settings.

\subsection{Cost}
At list prices of October 2026, running the 23{,}652 reviews costs about \$1{,}380, or 5.8 cents per review: Jev \$1 (2{,}700 reviews), Haiku \$171 (5{,}400), Sonnet \$702 (2{,}400), Flash \$66 (5{,}400), Pro \$190 (5{,}400) and Argon \$248 (2{,}352) at its introductory price, which doubles on 1 December 2026. Batch mode halves the price of the Claude and Gemini models, bringing the total to about \$690.

\section{Results}
\label{sec:results}

\subsection{Six reviewers, six views on the same papers}

\begin{figure}[t]
\centering
\includegraphics[width=\columnwidth]{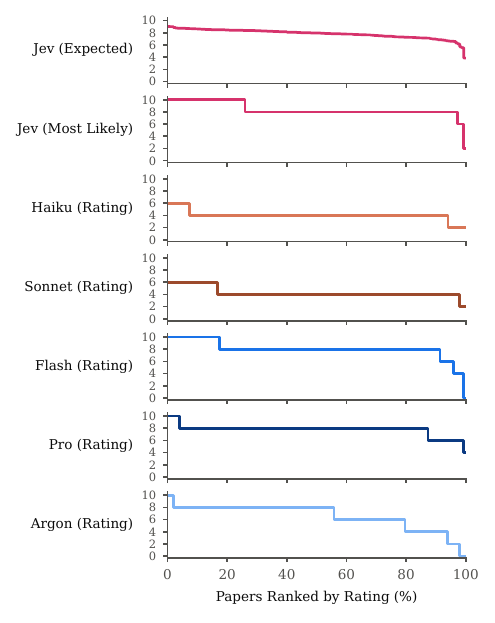}
\caption{Each reviewer's ratings of the 150 papers under \stmt{No AI}, on its first review, with papers ranked by rating. Jev with its expected and its most likely rating; the language models in the rating setting.}
\label{fig:ratings}
\end{figure}

\begin{figure}[t]
\centering
\includegraphics[width=\columnwidth]{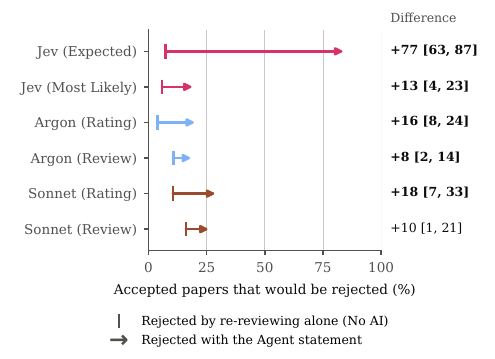}
\caption{Accepted papers that would be rejected, for the settings with a significant rating penalty: by re-reviewing alone under \stmt{No AI} and with the \stmt{Agent} statement, when the venue keeps the best 27.4\%, ICLR 2026's acceptance rate. On the right, the difference in percentage points with its 95\% CI; in bold if it passes the Holm correction.}
\label{fig:incentive}
\end{figure}

The reviewers judge the same papers very differently (Figure~\ref{fig:ratings}). Under the \stmt{No AI} statement, Jev (on its most likely rating), Flash and Pro are generous: they rate the papers at 8.43, 8.08 and 7.73 on average, close to accept, and rate 99, 96 and 99\% of them weak accept or higher. Haiku, in contrast, is stricter, rating the papers at 3.95, borderline reject, and rating only 7\% weak accept or higher. Argon lies in between, with an average rating of 6.60, and rates 80\% of the papers weak accept or higher. Sonnet, similar to Haiku, is also strict: it rates the papers at 4.29 and only 17\% weak accept or higher. Furthermore, the reviewers do not agree much on which papers are better: between models, the Spearman rank correlation of their ratings lies between 0.15 and 0.66.

ICLR 2026 accepted 27.4\% of valid submissions~\cite{iclr2026retrospective}. None of the reviewers comes close to that rate: Jev, Flash, Pro and Argon rate 80 to 99\% of the papers weak accept or higher, while Haiku and Sonnet do so for only 7 to 17\%.

Just as important is how consistent each reviewer is. Jev and Sonnet, asked for a rating alone, are the most consistent: their two reviews of a paper give a different rating in 5 and 6\% of the cases. The other settings do so in 10 to 38\%. On average, two reviews differ by 0.14 points for Jev (on its most likely rating) and by 0.12 to 0.81 for the language models.

\begin{table*}[t]
\caption{Rating difference between two statements: the first minus the second, on the 0--10 review scale, mean over 150 papers. Bold: significant after Holm's correction, two-sided, within the row (one family per comparison). Rating: the model is asked to answer the form directly; review: to write a review first, then answer the form. Jev: the expected value and the mode of the rating distribution. Sonnet and Argon review the papers under four statements only, to limit cost.}
\label{tab:judgment}
\centering
\small
\setlength{\tabcolsep}{3.5pt}
\begin{tabular}{lrrrrrrrrrrrr}
\toprule
 & \multicolumn{2}{c}{Jev} & \multicolumn{2}{c}{Haiku} & \multicolumn{2}{c}{Sonnet} & \multicolumn{2}{c}{Flash} & \multicolumn{2}{c}{Pro} & \multicolumn{2}{c}{Argon} \\
\cmidrule(lr){2-3}\cmidrule(lr){4-5}\cmidrule(lr){6-7}\cmidrule(lr){8-9}\cmidrule(lr){10-11}\cmidrule(lr){12-13}
Comparison & expected & mode & rating & review & rating & review & rating & review & rating & review & rating & review \\
\midrule
\stmt{Writing} vs.\ \stmt{No AI} & $+0.01$ & $-0.01$ & $+0.03$ & $+0.05$ & $-0.05$ & $+0.03$ & $+0.07$ & $+0.13$ & $-0.03$ & $-0.07$ & $-0.06$ & $-0.03$ \\
\stmt{Code} vs.\ \stmt{No AI} & $+0.02$ & $-0.01$ & $-0.02$ & $+0.06$ &  &  & $-0.03$ & $+0.09$ & $+0.02$ & $+0.08$ &  & \\
\stmt{Research help} vs.\ \stmt{No AI} & $-0.01$ & $-0.03$ & $+0.00$ & $-0.04$ &  &  & $-0.02$ & $+0.05$ & $-0.05$ & $-0.05$ &  & \\
\stmt{Agent} vs.\ \stmt{No AI} & $\mathbf{-0.44}$ & $\mathbf{-0.11}$ & $-0.13$ & $-0.04$ & $\mathbf{-0.19}$ & $\mathbf{-0.13}$ & $-0.14$ & $-0.03$ & $-0.09$ & $+0.03$ & $\mathbf{-0.48}$ & $\mathbf{-0.32}$ \\
\stmt{AI author} vs.\ \stmt{No AI} & $\mathbf{-2.68}$ & $\mathbf{-2.42}$ & $\mathbf{-0.45}$ & $\mathbf{-0.22}$ & $\mathbf{-0.37}$ & $\mathbf{-0.27}$ & $\mathbf{-4.12}$ & $\mathbf{-0.51}$ & $\mathbf{-0.42}$ & $\mathbf{-0.47}$ & $\mathbf{-0.52}$ & $\mathbf{-0.43}$ \\
\midrule
\stmt{Agent (ChatGPT)} vs.\ \stmt{No AI} & $\mathbf{-0.73}$ & $\mathbf{-0.13}$ & $\mathbf{-0.16}$ & $\mathbf{-0.15}$ &  &  & $\mathbf{-0.27}$ & $-0.10$ & $-0.13$ & $-0.13$ &  & \\
\stmt{Agent (Claude)} vs.\ \stmt{No AI} & $\mathbf{-0.32}$ & $\mathbf{-0.09}$ & $-0.07$ & $+0.03$ &  &  & $-0.19$ & $-0.03$ & $-0.04$ & $-0.05$ &  & \\
\stmt{Agent (Gemini)} vs.\ \stmt{No AI} & $\mathbf{-0.31}$ & $-0.07$ & $\mathbf{-0.17}$ & $-0.07$ &  &  & $-0.12$ & $-0.05$ & $-0.11$ & $-0.11$ &  & \\
\bottomrule
\end{tabular}
\end{table*}

\subsection{No evidence of a disclosure bias for assistance}

Table~\ref{tab:judgment} shows the main results. For writing help, AI-written code and research help, no reviewer's rating differs significantly from its rating under the \stmt{No AI} statement. This does not mean that the statements have no effect, but it does limit how large such an effect might be: the differences lie between $-0.07$ and $+0.13$ points, and no 95\% confidence interval reaches more than 0.21 points below no AI. Only the agent statement, for some reviewers, and the AI author statement, for all, lower the rating significantly. 

\subsection{Some evidence for a disclosure bias against AI agents and authors}
An AI author is clearly penalised by every reviewer in all settings, by 0.22 points (Haiku after a review) up to 4.12 (Flash, rating alone). The research agent statement lies between assistance and an AI author, and here three reviewers penalise it significantly. Against \stmt{No AI}, averaged over its two reviews of each paper, Jev's expected rating falls for 144 of the 150 papers, by 0.44 points on average, and its most likely rating by 0.11. Argon lowers its rating by 0.48 points when asked for a rating alone, lowering 44 papers and raising 8, and by 0.32 points after a written review. Sonnet lowers its rating by 0.19 points when asked for a rating alone, lowering 20 papers and raising 2, and by 0.13 points after a written review. Haiku and Flash, asked for a rating alone, lower theirs by 0.13 and 0.14 points, and Pro by 0.09; none of these three passes the correction. After a written review, Haiku, Flash and Pro show no significant penalty either ($-0.04$, $-0.03$ and $+0.03$).

With the agent's model named (Table~\ref{tab:judgment}, last three rows), Jev's expected rating is significantly lower than under \stmt{No AI} for all three named agents, and its most likely rating for those named ChatGPT and Claude. Haiku, asked for a rating alone, rates the agents named ChatGPT and Gemini significantly lower, and after a written review the one named ChatGPT. Flash, asked for a rating alone, rates the agent named ChatGPT significantly lower. Pro shows no significant penalty for any named model.

\subsection{From a rating penalty to a decision}
Figure~\ref{fig:incentive} shows how many of each reviewer's accepted papers the \stmt{Agent} statement rejects beyond re-reviewing alone, for the settings with a significant rating penalty. Jev, on its most likely rating, rejects 13 percentage points more of its accepted papers than re-reviewing alone (95\% CI 4 to 23). Argon rejects 16 percentage points more when asked for a rating alone (8 to 24), and 8 more after a written review (2 to 14). Sonnet rejects 18 percentage points more when asked for a rating alone (7 to 33), and 10 more after a written review (1 to 21). Of the effects in Figure~\ref{fig:incentive}, only Sonnet's after a written review does not pass the correction. Haiku, Flash and Pro show no significant rating penalty. For them, the statement rejects at most 4 percentage points more than re-reviewing alone, and none of these effects passes the correction.

The effect also depends on how the reviewer's output is used. Jev's expected rating, the hypothetical reading from Section~\ref{sec:method}, is lowered by the \stmt{Agent} statement for nearly every paper, by 0.44 points on average. In comparison, the most likely rating drops by 0.11 on average. The expected rating takes 111 distinct values, against at most six for the most likely rating, and many accepted papers lie just above the boundary. A small drop takes many of them below it. As a consequence, on the expected rating, the statement rejects 77 percentage points more than re-reviewing alone, against 13 for Jev's most likely rating (Figure~\ref{fig:incentive}). 

\begin{figure}[t]
\centering
\includegraphics[width=\columnwidth]{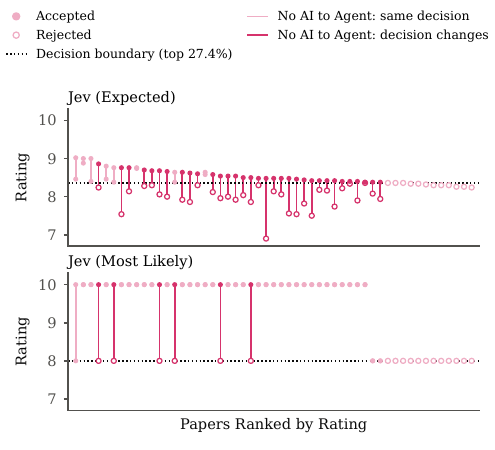}
\caption{Jev's ratings under \stmt{No AI} and, for the accepted papers, under the \stmt{Agent} statement, ranked by the rating under \stmt{No AI}. The decision changes for 33 of the 41 accepted papers by the expected rating and for 6 of the 41 by the most likely rating, for one order of the two reviews and one tie-break. Figure~\ref{fig:incentive} pools both orders and subtracts re-reviewing alone.}
\label{fig:shift}
\end{figure}

Figure~\ref{fig:shift} shows both readings of Jev for one assignment of reviews. By the most likely rating, most ratings do not move, and a paper that falls lands on the boundary, tied with the rejected papers. For Jev, the statement moves probability from the top levels to all lower ones, so the expected rating falls while the top level often remains the most likely.

\section{Discussion}
\label{sec:discussion}
\paragraph{A price on honesty?} Disclosure policies assume that an honest statement does not hurt a sound paper. For the AI reviewers we studied, we found no evidence of a disclosure bias as long as the AI only assisted. Once the statement says that an AI agent planned and ran the experiments, the assumption holds only for some reviewers: Jev, Sonnet and Argon rate the paper lower than under \stmt{No AI}. Once it says that AI did all the work, every reviewer rates the paper lower. Whether an AI reviewer penalises the agent statement thus has to be measured for each reviewer. 

One may object that a reviewer has reason to doubt experiments that an AI planned and ran. Yet our AI reviewers apply that doubt without checking anything (such as the code or the data), to a paper whose experiments they have just read. Doubt can be justified, but lowering the rating for the disclosure alone reflects an inclination or bias against certain uses of AI rather than a fair judgement of the work. Doubt should instead lead a reviewer to scrutinise the work itself, its evidence, code and data. Since the papers in our experiment are identical except for their AI-use statement, the discount can only come from what the reviewer reads about the AI's part in the research. This raises a question: is it sound for any reviewer, human or AI, to judge a paper by who or what tool was part of the research process? Peer review is meant to judge the content of a paper, and double-blind review hides the authors so that their identity cannot influence a reviewer's judgement. The same reasoning may apply to AI-use statements: they tell a reviewer which tools took part, and can sway the judgement just as an author's identity can. Disclosure serves transparency and accountability; it is not a measure of a paper's quality. Should a reviewer rate a paper less because a master's student collected the data, a PhD student wrote the code or a postdoc ran the analysis, rather than the principal investigator? A reviewer should probably judge the data, the code and the analysis, not who produced them. For authors facing a reviewer, human or AI, that penalises disclosed AI use, the incentive is clear: understate the AI's role. Declared AI use already lags behind actual use~\cite{he2025policies}, and a disclosure bias only widens this gap, precisely for the agentic work that we would most like to see reported.

\paragraph{What turns a penalty into a rejection.} Whether a rating penalty rejects a paper depends on how the reviewer's output is used. A small shift in Jev's probabilities often leaves its most likely rating unchanged, while its expected rating moves with every shift.

The effect also depends on how many other papers disclose. When the venue keeps a fixed share of papers, only a paper's relative position counts. If every paper disclosed and lost the same number of points, the ranking would not change. Our reviewers, however, lower some papers more than others, so rankings could change even if every paper disclosed. We measure the simplest case: one author reporting agentic use among papers that report no AI use. It is somewhat realistic as long as few papers disclose agentic work: of the 160 statements we read, 24 do. What happens as more authors disclose, we did not test.

\paragraph{Implications.} Studies like ours can read as proof that AI reviewers are flawed and that reviewing should stay a purely human affair. We somewhat disagree. We did not compare our AI reviewers with human ones, and people may be just as biased and inconsistent: in a large experiment at NeurIPS 2021, 882 submissions were each reviewed by two independent committees, and of the papers one committee accepted, the other rejected about half~\cite{beygelzimer2023consistency}. The question is therefore not whether AI reviewers are flawless, but where they can improve a flawed process. For venues that use an AI reviewer to rate papers, our results suggest two things. First, test the reviewer on the decisions it will make, in the way its output will be used: the average rating penalty does not show how many papers it rejects, and the same reviews can cost few papers or many depending on how they are read. Second, consider not showing the reviewer the AI-use statement; venues can still require it without passing it on. We see the most room for AI reviewers elsewhere, in the objective, tedious work that a system under pressure cannot do today: checking a paper's claims, reproducing its results and finding mistakes in its references. For such checks, how the authors used AI should not matter. How interesting, novel or important a paper is, we would rather leave to people.

\section{Limitations}
\label{sec:limitations}

\paragraph{Models.} We study six reviewers: a decision model, a small and a larger language model from each of two providers, and Google's newest model. Among the language models, Sonnet and Argon penalise the \stmt{Agent} statement significantly, while Haiku, Flash and Pro do not, although Haiku and Flash do penalise some statements that name the agent's model. The bias thus does not seem to be a property of a single provider. None of our reviewers comes from OpenAI.

\paragraph{Reasoning effort.} The language models ran at low reasoning effort. Whether more reasoning reduces the disclosure bias against agents, or whether it makes a model more attentive to the statement, we did not test.

\paragraph{Design.} All 150 papers come from one venue, ICLR 2027. We simulate decisions with a single AI reviewer, while real decisions combine several reviews and an area chair. A growing share of those reviews is itself written with LLM help~\cite{sharma2026favor,kim2026icml}.

\paragraph{Statements.} We use one wording per level. Like the real statements they follow, neighbouring wordings differ in more than the AI's role, so we know which statements are penalised, but not which part of each causes the penalty. For example, the \stmt{AI author} statement is the only statement of AI use that does not say the authors take responsibility for the content. Our \stmt{Agent} statement is also plainer than most real ones. And we replaced only the AI-use section. Other parts of a paper may still mention AI use, and for a few papers the reviewers noticed the discrepancy.

\paragraph{Future work.} The most direct follow-up would vary one part of the \stmt{Agent} statement at a time, such as who acts or how the authors checked the work, with several wordings per level, to find out what the reviewers react to. Testing reviewers from OpenAI, higher reasoning effort and human reviewers on the same papers would then show how general the bias is.

\section{Conclusion}
\label{sec:conclusion}

We found no evidence of a disclosure bias for AI assistance, whereas some AI reviewers penalise the disclosure that an AI agent ran the experiments under the authors' review, and all penalise the disclosure that AI did all the work. How much this costs a paper depends on how a reviewer's ratings become decisions, and may depend on how many papers disclose. As more papers are both produced with AI research agents and reviewed by AI, venues that require disclosure should therefore test whether changing the statement changes acceptance decisions under their own review procedure.

\section*{AI Use Disclosure}
We used Anthropic's Claude Opus~5.5 through Claude Code. The core ideas and design decisions came from the authors. Claude helped design the experiments, wrote the code, assisted with the analysis, searched for relevant literature and drafted the text. The authors reviewed every step and take responsibility for the content.

\bibliography{ref}

\end{document}